\documentclass[letterpaper]{article} 
\usepackage[preprint]{aaai2027}  
\usepackage[hyphens]{url}  
\usepackage{graphicx} 
\usepackage{natbib}  
\usepackage{caption} 
\usepackage{algorithm}
\usepackage{algorithmic}

\usepackage{newfloat}
\usepackage{listings}
\DeclareCaptionStyle{ruled}{labelfont=normalfont,labelsep=colon,strut=off} 
\floatstyle{ruled}
\newfloat{listing}{tb}{lst}{}
\floatname{listing}{Listing}

\usepackage{booktabs}
\usepackage{amsmath,amssymb}
\usepackage{multirow}
\title{VoxStruct3D: Structure-Leading Flow Matching for Voxel-Space 3D MRI Synthesis}
\author{
    Fang Li\textsuperscript{\rm 1},
    Yang Gao\textsuperscript{\rm 1}\corresponding,
    Shihao Zou\textsuperscript{\rm 2},
    Weixin Si\textsuperscript{\rm 3},
    Hongyu Wu\textsuperscript{\rm 1},
    Qing Xia\textsuperscript{\rm 1},
    Shuai Li\textsuperscript{\rm 1},
    Aimin Hao\textsuperscript{\rm 1}
}
\affiliations{
    \textsuperscript{\rm 1} State Key Laboratory of Virtual Reality Technology and Systems, Beihang University\\
    \textsuperscript{\rm 2} Shenzhen Institutes of Advanced Technology, Chinese Academy of Sciences\\
    \textsuperscript{\rm 3} Faculty of Computer Science and Control Engineering, Shenzhen University of Advanced Technology\\
    
    \tt\small {neesky, gaoyangvr}@buaa.edu.cn
}

\begin{document}

\maketitle

\begin{abstract}
High-fidelity 3D MRI synthesis requires both globally coherent anatomy and fine-grained voxel-level detail. Although latent diffusion makes volumetric generation tractable, its image autoencoder introduces a reconstruction bottleneck that can limit the fine detail recoverable in the final volume. We present VoxStruct3D, a voxel-space flow-matching framework that directly models full-resolution MRI volumes using a clean-data prediction objective. Its Volumetric Voxel Generator (VVG) combines factorized 3D patch embedding with overlapping upsampling, time-modulated residual refinement, and skip fusion, enabling neighboring tokens to jointly reconstruct shared voxel regions and suppress patch-boundary artifacts. To complement direct voxel-space modeling with an explicit anatomical prior, we further introduce a Structure-First, Image-Follows (SFIF) strategy. A frozen pretrained 3D medical encoder and a StructVAE extract compact structure tokens that preserve dominant anatomy, while a structure-leading schedule keeps their trajectory ahead of the image trajectory. Patch-Aligned RoPE spatially aligns the unequal token grids, and asymmetric attention enforces one-way guidance from structure to image.
Experiments on pathological and healthy T1-weighted brain MRI datasets show that VoxStruct3D achieves the strongest overall performance across feature-distribution alignment, sample diversity, and perceptual quality, producing anatomically coherent and visually realistic volumes.

\end{abstract}

\section{Introduction}
\label{sec:intro}

High-fidelity 3D MRI synthesis has been widely studied to support medical-image analysis tasks such as diagnosis, segmentation, anomaly detection, and treatment planning \citep{kazerouni2022diffusion,dorjsembe2024conditional,xu2025_3dino}. Existing generative models, however, still struggle to balance local detail fidelity with globally coherent anatomy. Such errors are particularly consequential in medical volumes, which tolerate less structural distortion than natural images: structural deformations and cross-slice inconsistencies can compromise the plausibility of the entire volume, while blurred boundaries and missing local details can limit its utility for downstream analysis. Simultaneously preserving global anatomical structure and fine local detail therefore remains a central challenge in high-fidelity 3D MRI generation \citep{khader2023denoising}.

Many recent 3D medical diffusion models perform generation in a compressed latent space. A learned encoder first maps the volume to a compact representation, diffusion is performed in that space, and a decoder reconstructs the final image \citep{guo2025maisi,wang2025meddiffusion,xu2026super,zhao2026maisi}. This design substantially reduces the difficulty of volumetric generation, but it also places an image autoencoder in the final image-formation path. Anatomical edges and local details attenuated during encoding and decoding may not be recovered by subsequent latent denoising.

Recent pixel-space generators such as JiT and PixelDiT have achieved strong results in natural-image generation \citep{li2026jit,yu2026pixeldit}. In particular, JiT uses clean-data prediction to learn pixel generation directly, without relying on a pretrained image tokenizer. This provides a promising starting point for 3D medical-image generation. A direct extension, however, cannot simply apply 3D patchify and unpatchify around patch-level DiT modeling. In a plain JiT-style output head, each token independently reconstructs a non-overlapping block. Although the DiT captures global interactions among tokens, continuity across independently decoded neighboring blocks must still be learned implicitly. This local reconstruction problem is more pronounced in 3D: adjacent 2D patches meet along 1D edges, whereas adjacent volumetric patches meet along entire 2D faces. This can consequently lead to conspicuous grid-aligned block artifacts, as shown in the left of Fig.~\ref{fig:qualitative_ablation}.

Meanwhile, the image autoencoder in latent diffusion contributes more than computational compression. Its learned latent representation can retain coarse semantic and spatial organization, providing a compact state from which global structure can be modeled \citep{ldm,pinaya2022brain}. Removing the image autoencoder also removes this compact structural abstraction, forcing a voxel-space model to infer global anatomy directly from a high-dimensional corrupted volume. This challenge is further amplified in 3D, where the additional depth dimension substantially increases the scale of spatial modeling and the complexity of long-range anatomical dependencies. Direct voxel-space generation can therefore occasionally produce anatomically implausible structures, as illustrated in the middle panel of Fig.~\ref{fig:qualitative_ablation}.

To address these two problems, we propose \textbf{VoxStruct3D}, a voxel-space flow-matching framework for high-fidelity 3D medical-image generation. First, we introduce the \emph{Volumetric Voxel Generator} (VVG) to eliminate grid-aligned block artifacts. VVG uses a DiT backbone to model global interactions among patch tokens, while replacing independent unpatchify with an overlapping volumetric decoder that allows neighboring tokens to jointly reconstruct shared voxel regions. A U-Net-style skip connection further carries fine-grained encoder features into the decoder. VVG thereby combines global token modeling with locally coupled reconstruction across adjacent voxel blocks. Second, to restore the missing structural guidance, we introduce \emph{Structure First, Image Follows} (SFIF). SFIF jointly generates a compact anatomical state distilled from a frozen pretrained 3DINO encoder \citep{xu2025_3dino}. Structure-leading clocks offset the flow-matching time steps so that structural information remains ahead of image formation throughout generation. Patch-Aligned RoPE (PA-RoPE) and an asymmetric attention mask then allow the image stream to read this structural state while blocking the reverse information path. The resulting structure state serves only as an internal anatomical guide and the output volume is generated entirely in the original voxel domain. 

Our contributions are:
\begin{itemize}
    \item We present VoxStruct3D, a voxel-space framework for synthesizing anatomically coherent, detail-preserving 3D MRI volumes, providing a scalable source of high-quality volumetric data to support the development of downstream clinical AI systems.
    \item We develop VVG to suppress block artifacts and preserve fine details through overlapping volumetric decoding, and SFIF to generate a compact 3DINO-derived anatomical state ahead of the image stream for aligned, one-way guidance via PA-RoPE and asymmetric attention.
    \item Experiments on pathological and healthy brain MRI demonstrate the effectiveness of VoxStruct3D, with ablations validating the distinct roles of VVG and SFIF.
\end{itemize}

\section{Related Work}
\label{sec:related_work}

\paragraph{Pixel-Space Generation in Natural and Medical Imaging.}
Recent natural-image generators have revisited direct pixel modeling. PixelFlow removes the pretrained VAE\cite{pixelflow}, JiT enables raw-patch generation through clean-data prediction\cite{li2026jit}, PixelDiT separates global token modeling from local detail recovery\cite{yu2026pixeldit}, and PiD replaces deterministic latent decoding with conditional pixel diffusion~\citep{lu2026pid}. However, these methods are mainly designed for 2D images. In particular, directly extending JiT's independent token-to-patch projection to 3D produces large voxel blocks without local blending, making discontinuities more likely along patch boundaries. VoxStruct3D instead combines clean-volume prediction, factorized 3D patch embedding, and overlapping volumetric reconstruction.

Medical studies have also explored direct volume generation. Med-DDPM \cite{dorjsembe2024conditional}extends noise-predicting U-Nets to 3D MRI, while many existing methods perform generation in a fixed 3D wavelet space~\citep{friedrich2024wdm,danese2026flowlet,tur2026wfm}. These approaches demonstrate the feasibility of volume-domain generation but retain conventional 3D denoisers or fixed transforms. In contrast, VoxStruct3D addresses the Transformer-specific challenges of efficiently tokenizing raw volumes, avoiding inter-block artifacts, and maintaining coherent anatomy under severe pixel corruption.

\paragraph{Structural Priors from Pretrained Encoders.}
Pretrained visual encoders are increasingly used to guide diffusion models. DINOv2 ~\citep{oquab2024dinov2,simeoni2025dinov3} provides transferable representations, while REPA and U-REPA align generative features with clean-image encoder features~\citep{repa,tian2026u}. However, such methods depend on compatible spatial grids and mainly provide training-time feature supervision.

Medical pretraining has similarly advanced from Med3D to volumetric encoders such as 3DINO~\citep{chen2019med3d,xu2025_3dino}. Direct tokenwise alignment remains difficult when the encoder and generator use different 3D grids. SFIF instead preserves descriptors on their native grid, converts them into a separately generated structure state, and uses PA-RoPE and asymmetric attention for cross-grid interaction. This allows the image stream to follow a progressively refined structural trajectory while preventing reverse leakage of local appearance.

\begin{figure*}[t]
    \centering
    \includegraphics[width=0.9\textwidth]{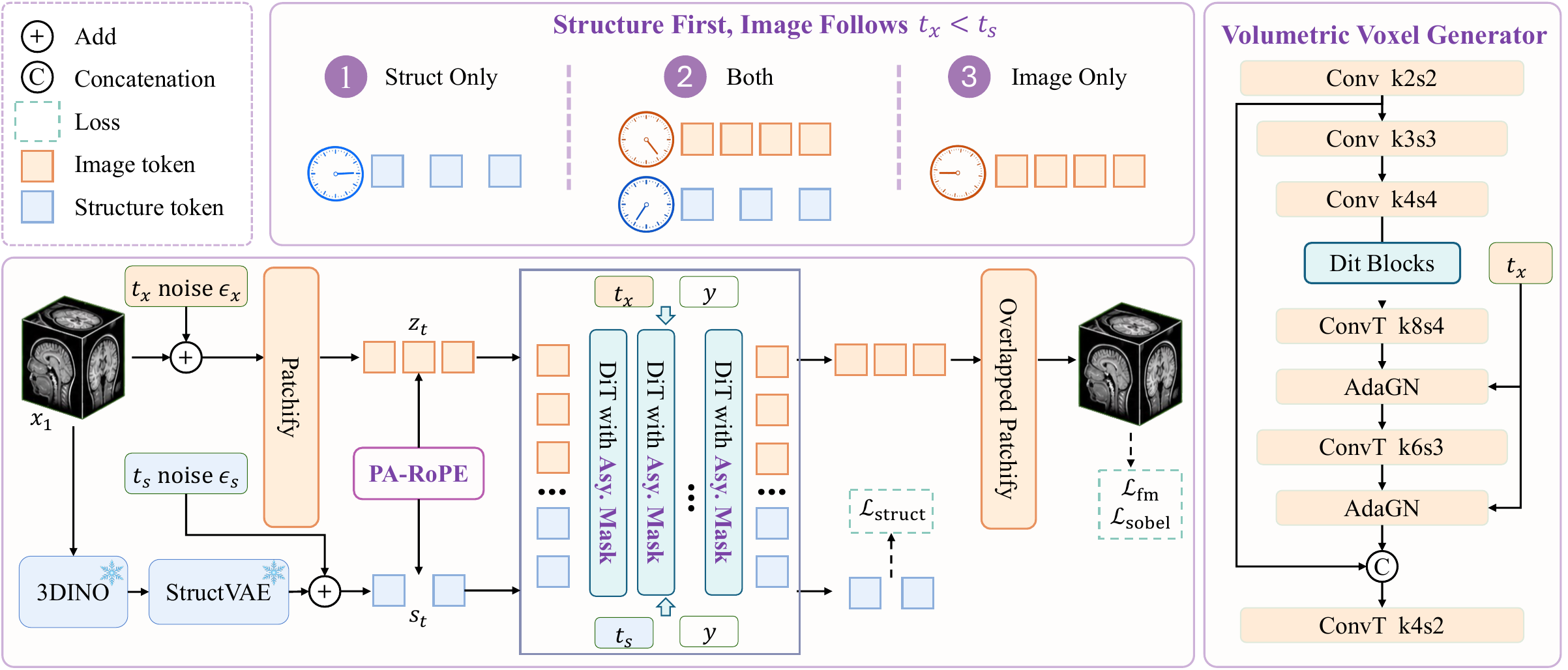}
    \caption{\textbf{Overview of Stage II}. Given a clean volume $\mathbf{x}_1$ and class label $y$, independent noise is injected into the image and structure streams at times $t_x$ and $t_s$, respectively. The noised volume is patchified into image tokens $\mathbf{z}_t$, while the frozen 3DINO encoder and StructVAE produce noised structure tokens $\mathbf{s}_t$. Both token streams are augmented with Patch-Aligned RoPE and jointly processed by a shared dual-stream DiT under an asymmetric attention mask. The structure branch is supervised in token space, whereas the image branch is reconstructed into the voxel domain through an overlapping unpatchify module and supervised against $\mathbf{x}_1$.}
    \label{fig:framework}
\end{figure*}

\section{Method}
\label{sec:method}

\paragraph{Overview.}
The training of VoxStruct3D consists of two stages. In Stage I, we train only the StructVAE, which compresses features extracted by a frozen 3DINO encoder\cite{xu2025_3dino} into compact structure tokens, as detailed in the SFIF subsection. Stage II follows the pipeline illustrated in Fig.~\ref{fig:framework}. Given a clean volume $\mathbf{x}_1 \in \mathbb{R}^{C \times D \times H \times W}$ and its class label $y$, where $D$, $H$, and $W$ denote the spatial depth, height, and width, respectively, and $C$ denotes the number of channels, we inject voxel-level noise at image time $t_x$ to obtain $\mathbf{x}_t$. The noised volume is then patchified into image tokens $\mathbf{z}_t \in \mathbb{R}^{N_{\mathrm{img}} \times d}$,
where $N_{\mathrm{img}}$ is the number of image tokens and $d$ is the token dimension. In parallel, the frozen 3DINO encoder and StructVAE extract clean structure tokens $\mathbf{s}_1 \in \mathbb{R}^{N_{\mathrm{str}} \times d}$. Independent noise is injected at structure time $t_s$ to produce $\mathbf{s}_t \in \mathbb{R}^{N_{\mathrm{str}} \times d}$, where $N_{\mathrm{str}}$ denotes the number of structure tokens. The image and structure tokens are then augmented with Patch-Aligned RoPE and jointly processed by a shared dual-stream DiT under an asymmetric attention mask.  The structure branch is supervised directly in token space, whereas the image branch is reconstructed into the voxel domain through an overlapping unpatchify module and supervised against the clean volume. Specifically, we sample two independent Gaussian noises, $\boldsymbol{\epsilon}_x,\boldsymbol{\epsilon}_s \sim \mathcal{N}(\mathbf{0},\mathbf{I})$, and define the two interpolation paths as
\begin{equation}
    \mathbf{x}_t
    =
    t_x\mathbf{x}_1
    +
    (1-t_x)\boldsymbol{\epsilon}_x,
    \qquad
    \mathbf{s}_t
    =
    t_s\mathbf{s}_1
    +
    (1-t_s)\boldsymbol{\epsilon}_s.
    \label{eq:flow_path}
\end{equation}
The model jointly predicts the clean endpoints $\mathbf{x}_1$ and $\mathbf{s}_1$. At inference, only the image branch generates and returns a volume in the original voxel domain, while the structure tokens act solely as an internal anatomical guide.

\subsection{Volumetric Voxel Generator}
\label{sec:VVG}

\paragraph{Shared Dual-stream DiT.}
This module facilitates patch-level information interaction between the image and structure streams. Specifically, the image tokens $z_{t}$ and structure tokens $s_{t}$ are jointly processed by a shared stack of DiT blocks. They correspond to different diffusion timesteps and therefore require stream-specific condition injection. The conditioning embedding is generally formulated as
\begin{equation}
    \mathbf{c}=\mathcal{T}(t)+\mathcal{Y}(y),
    \label{eq:image_condition}
\end{equation}
where $\mathcal{T}$ and $\mathcal{Y}$ denote the timestep and class embedding functions, respectively. Accordingly, the image stream is conditioned by
$\mathbf{c}_z=\mathcal{T}(t_x)+\mathcal{Y}(y)$,
whereas the structure stream uses
$\mathbf{c}_s=\mathcal{T}(t_s)+\mathcal{Y}(y)$.
The two conditions are independently injected into their corresponding token streams through AdaLN-Zero, allowing the shared DiT blocks to adapt their feature modulation to different noise levels while preserving cross-stream interaction.

In addition, we append 32 learnable context tokens to the joint token sequence. These tokens act as class context carriers and attention sinks~\citep{xiao2024streamingllm}, facilitating information aggregation and communication between the two streams. Class-label dropout is further applied during training to enable classifier-free guidance.

\paragraph{Overlapping Unpatchify.}
A conventional unpatchify module independently maps each output token to a non-overlapping volumetric block. As discussed in the introduction, although the Transformer enables global information exchange, this independent final reconstruction may still produce visible seams between adjacent blocks.

To alleviate this issue, we replace independent block recovery with a lightweight overlapping volumetric decoder. The image tokens are first rearranged into their corresponding 3D grid and then progressively upsampled using three transposed-convolution stages with kernel--stride pairs
$(k,s)=(8,4)$, $(6,3)$, and $(4,2)$.
Because each kernel is larger than its corresponding stride, adjacent tokens contribute to overlapping voxel regions rather than reconstructing isolated blocks. The overlapping predictions are therefore locally fused during upsampling, substantially reducing grid-like discontinuities at patch boundaries.

Each transposed-convolution stage is followed by a linear feature projection modulated by the image timestep $t_x$, enabling timestep-aware refinement throughout the reconstruction process. The patchify encoder mirrors the unpatchify decoder using 3D convolutional stages with the corresponding kernel sizes and strides. Features at matched resolutions are connected through U-Net-style skip connections, which preserve fine-grained spatial information and improve volumetric reconstruction. This design effectively weakens block artifacts without relying on an additional image autoencoder.

\subsection{Structure First, Image Follows}
\label{sec:sfif}

\paragraph{StructVAE.}
We use a frozen 3DINO ViT-L/16 \citep{xu2025_3dino} as the structural teacher. Its features contain substantially richer semantic and appearance information than is required to guide pixel generation. Directly treating these high-dimensional descriptors as a generative target would therefore introduce an unnecessarily difficult structure trajectory. Stage I instead trains structVAE to compress the teacher features and retain the dominant anatomical information needed by the image stream:
\begin{equation}
    \begin{aligned}
        \mathbf{r}_1
        &=f_{\mathrm{3DINO}}(\mathbf{x}_1)
        \in\mathbb{R}^{N_{\mathrm{str}}\times d},\\
        \mathbf{s}_1
        &=\mathbf{W}_{\mathrm{code}}\boldsymbol{\mu}_{\psi}(\mathbf{r}_1)
        \in\mathbb{R}^{N_{\mathrm{str}}\times d}.
    \end{aligned}
    \label{eq:structure_tokens}
\end{equation}
Here, $\mathbf{x}_1$ is the clean volume, $f_{\mathrm{3DINO}}$ is the frozen teacher, and $\mathbf{r}_1$ contains its patch features. The structVAE encoder with parameters $\psi$ predicts posterior mean $\boldsymbol{\mu}_{\psi}$, and $\mathbf{W}_{\mathrm{code}}$ projects it into the structure-token space to produce $\mathbf{s}_1$. During Stage I, a lightweight decoder reconstructs $\widehat{\mathbf{r}}_1$ from the sampled structure code and compares it with $\mathbf{r}_1$. This reconstruction bottleneck makes structVAE function as a learned low-pass filter: it discards incidental appearance variation while retaining the dominant components required to reconstruct the teacher features. The complete Stage-I objective is given in Eq.~\ref{eq:stage1_loss}. In Stage II, we use the posterior mean to construct deterministic structure tokens and freeze both 3DINO and structVAE.

\paragraph{Structure-leading clocks.}
Because $\mathbf{s}_1$ is a lossy summary that retains only dominant anatomy, reconstructing the structure state is simpler than reconstructing the full voxel volume. One direct strategy would first generate the complete structure state and then run a second generative process conditioned on it. This serial design, however, requires two separate generation passes. The image stream also need not wait for a fully resolved structure state; it can benefit from guidance that becomes progressively clearer. We therefore generate both streams within one DiT trajectory while advancing the structure stream ahead of the image stream.

Let $t\in[0,1]$ follow the flow-matching time distribution, let $\delta>0$ denote the structure lead, and draw $b\sim\operatorname{Bernoulli}(p_{\mathrm{warm}})$. We sample the paired clocks as
\begin{equation}
    (t_x,t_s)=
    \begin{cases}
        (t,\min(t+\delta,1)),
        & b=0,\\
        (0,u),\quad u\sim\mathcal{U}(0,\delta),
        & b=1.
    \end{cases}
    \label{eq:paired_clocks}
\end{equation}
Thus, $t_s\geq t_x$ and the structure state is always at least as close to its clean endpoint as the image state. The second branch provides structure-only samples that cover the initial structure-formation interval. Independent Gaussian noise at $t_x$ and $t_s$ produces $\mathbf{x}_t$ and $\mathbf{s}_t$, respectively. The structure loss is applied to every sample, whereas the image loss is masked for structure-only samples.

\begin{table*}[t]
\centering

{%
\fontsize{9pt}{10.8pt}\selectfont
\rmfamily
\setlength{\tabcolsep}{1mm}
\renewcommand{\arraystretch}{1.0}

\begin{tabular}{llcccccccc}
\toprule

Dataset & Method & \multicolumn{2}{c}{MedicalNet ($\times 10^3$)} & \multicolumn{2}{c}{3DINO} & MS-SSIM$\downarrow$ & MUSIQ$\uparrow$ & NIQE$\downarrow$ & Sharpness$\uparrow$ \\
\cmidrule(lr){3-4}\cmidrule(lr){5-6}
 & & FID$\downarrow$ & MMD-RBF$\downarrow$ & FID$\downarrow$ & MMD-RBF$\downarrow$ & & & & 3D Tenengrad \\
\midrule
\multirow[c]{7}{*}{Pathological}
 & Real &  &  &  &  & 0.7455 & 44.03 & 9.025 & 2.320 \\
 & HA-GAN & 3.814 & 40.26 & 83.08 & 1.417 & 0.9696 & 14.55 & 36.08 & 4.230 \\
 & 3D-LDM & 12.84 & 68.73 & 112.6 & 1.621 & 0.9825 & 21.47 & 27.36 & 1.742 \\
 & 3D MedDiffusion & 0.4047 & 3.781 & 17.51 & 0.6565 & 0.8023 & 38.78 & 8.273 & 1.955 \\
 & WDM & 0.3621 & 3.614 & 16.84 & 0.5734 & 0.7938 & 39.26 & 8.014 & 2.034 \\
 & MOTFM & 1.100 & 18.56 & 42.87 & 1.192 & 0.9090 & 25.83 & 23.01 & 1.696 \\
 & VoxStruct3D (Ours) & 0.3365 & 3.297 & 12.89 & 0.1600 & 0.7413 & 39.38 & 7.703 & 2.159 \\

\midrule
\multirow[c]{7}{*}{Healthy}
 & Real &  &  &  &  & 0.7240 & 46.60 & 12.83 & 2.707 \\
 & HA-GAN & 4.672 & 35.84 & 72.36 & 1.084 & 0.9517 & 18.12 & 36.74 & 5.146 \\
 & 3D-LDM & 12.37 & 61.25 & 101.4 & 1.376 & 0.9784 & 20.86 & 26.42 & 1.684 \\
 & 3D MedDiffusion & 1.893 & 7.146 & 11.08 & 0.1903 & 0.7915 & 35.82 & 9.864 & 2.012 \\
 & WDM & 1.746 & 6.824 & 10.32 & 0.1847 & 0.7828 & 36.41 & 9.537 & 2.054 \\
 & MOTFM & 3.426 & 24.73 & 39.64 & 0.8694 & 0.9136 & 26.71 & 21.38 & 1.858 \\
 & VoxStruct3D (Ours) & 1.572 & 6.107 & 8.476 & 0.0883 & 0.7320 & 37.06 & 9.045 & 2.236 \\
\bottomrule
\end{tabular}}
\caption{\textbf{Main comparison on pathological and healthy T1 brain MRI.} MedicalNet FID and MMD-RBF are scaled by $10^3$. Each method generates 1,000 volumes per condition, evaluated against 1,251 BraTS 2021 and 958 healthy real volumes. Real rows report reference values from the complete real datasets.}
\label{tab:main_results}
\end{table*}

\begin{figure}[!h]
    \centering
    \includegraphics[width=\columnwidth]{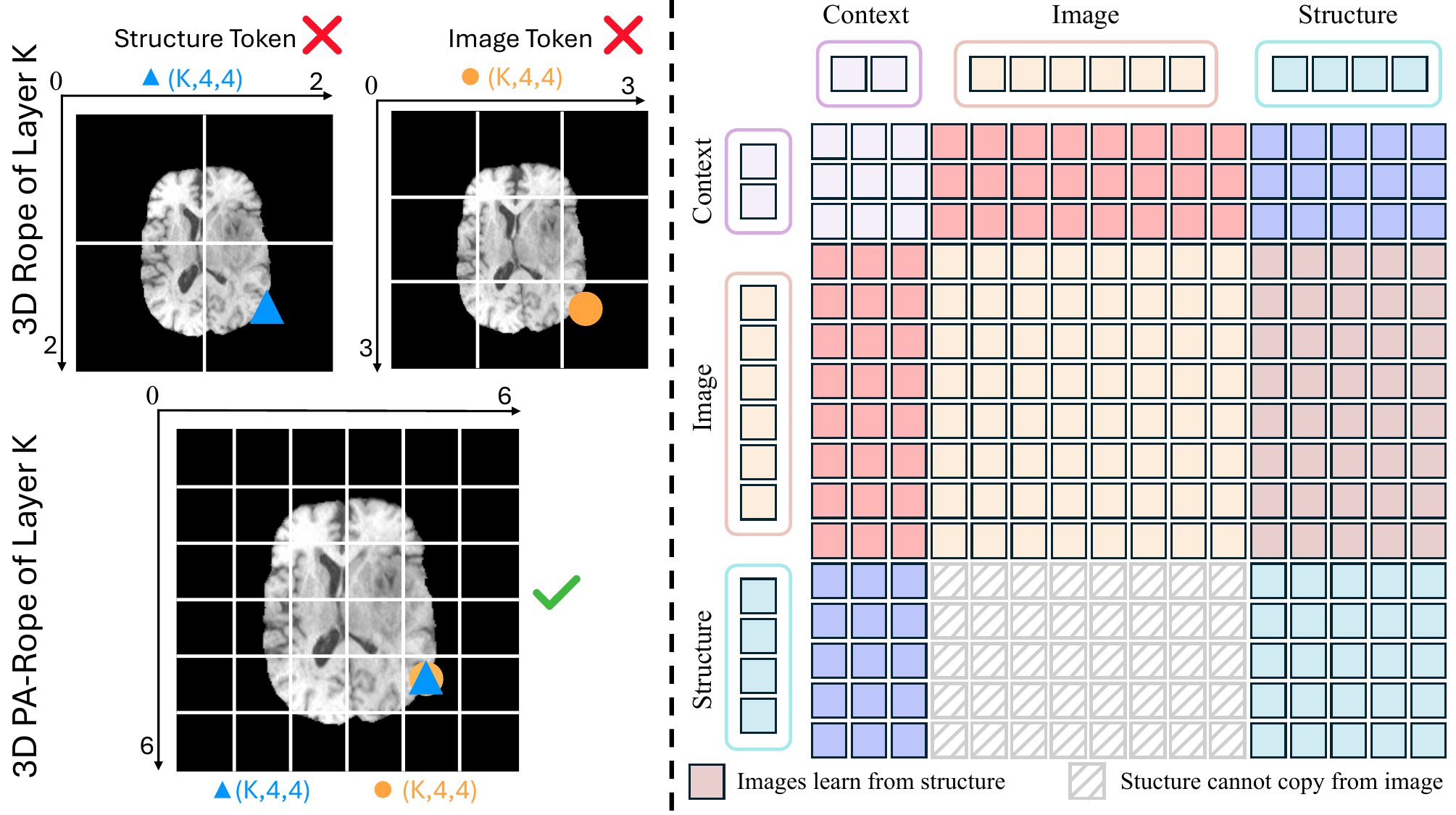}
    \caption{\textbf{PA-RoPE and asymmetric attention mask.} Left: image and structure tokens are mapped to a common spatial lattice. Right: image tokens can read structure tokens while the reverse path is blocked.}
    \label{fig:rope_attention}
\end{figure}

\paragraph{Asymmetric attention.}
The temporal lead determines which stream develops first, while an asymmetric attention mask determines how information flows between them. Image queries may attend to structure keys so that the progressively clarified anatomy can guide voxel generation. Structure queries are prevented from attending to image keys because the less advanced image state provides little useful structural evidence and may leak local appearance into the structure stream. Both streams retain access to the shared context token. The mask is illustrated on the right of Fig.~\ref{fig:rope_attention}.

\paragraph{Patch-Aligned RoPE.}
The image and 3DINO structure tokens occupy grids of different sizes, so applying 3D RoPE independently on their native integer coordinates would misalign their spatial positions during cross-stream attention. We address this problem with Patch-Aligned RoPE (PA-RoPE). As illustrated in 2D on the left of Fig.~\ref{fig:rope_attention}, we first reshape $\mathbf{z}_t$ and $\mathbf{s}_t$ into their spatial grids. Let $G_a^q$ denote the grid size of stream $q\in\{\mathrm{img},\mathrm{str}\}$ along axis $a\in\{D,H,W\}$. We construct a common lattice on each axis and map every token center onto it:
\begin{equation}
    L_a=\operatorname{lcm}(G_a^{\mathrm{img}},G_a^{\mathrm{str}}),
    \qquad
    \widetilde{g}_a^q
    =
    \frac{(2g_a^q+1)L_a}{2G_a^q}.
    \label{eq:pa_rope}
\end{equation}
Here, $g_a^q\in\{0,\ldots,G_a^q-1\}$ is the native token index. The common coordinates $\widetilde{g}_a^q$ are then used to compute the rotary vectors for both streams. Each attention head is split into depth, height, and width groups, with RoPE applied independently along each axis.
PA-RoPE therefore aligns image and structure tokens in a shared 3D coordinate system without interpolating either feature grid, and applies to arbitrary pairs of unequal grid sizes.

\paragraph{Joint generation.}
At inference, the same clocks produce a single coupled trajectory: the structure stream evolves alone during the initial $\delta$ interval, both streams then evolve jointly with structure remaining ahead, and the completed structure is finally held fixed while the image stream finishes. This schedule progressively injects anatomical guidance without requiring a separate structure-generation pass.

\subsection{Overall Objectives}
\label{sec:objectives}

Stage I trains structVAE while keeping 3DINO frozen. Its objective combines descriptor reconstruction, cosine consistency, KL regularization, and code-statistics regularization:
\begin{equation}
    \mathcal{L}_{\mathrm{I}}
    =\lambda_{\mathrm{rec}}\mathcal{L}_{\mathrm{rec}}
    +\lambda_{\mathrm{cos}}\mathcal{L}_{\mathrm{cos}}
    +\lambda_{\mathrm{KL}}\mathcal{L}_{\mathrm{KL}}
    +\lambda_{\mathrm{stat}}\mathcal{L}_{\mathrm{stat}}.
    \label{eq:stage1_loss}
\end{equation}

Following JiT \citep{li2026jit}, Stage II adopts $x$-prediction for the clean image and structure endpoints, while optimizing the two branches with velocity-space losses $\mathcal{L}_{\mathrm{image}}^v$ and $\mathcal{L}_{\mathrm{structure}}^v$. We additionally apply a time-weighted $\mathcal{L}_{\mathrm{Sobel}}$ to the clean-image prediction to encourage consistent local 3D gradients. The complete Stage-II objective is
\begin{equation}
    \mathcal{L}_{\mathrm{II}}
    =\mathcal{L}_{\mathrm{image}}^v
    +\mathcal{L}_{\mathrm{structure}}^v
    +\lambda_{\mathrm{sobel}}\mathcal{L}_{\mathrm{sobel}}.
    \label{eq:total_loss}
\end{equation}

\section{Experiments}
\label{sec:experiments}

\begin{figure*}[t]
\centering

    \includegraphics[width=0.9\textwidth]{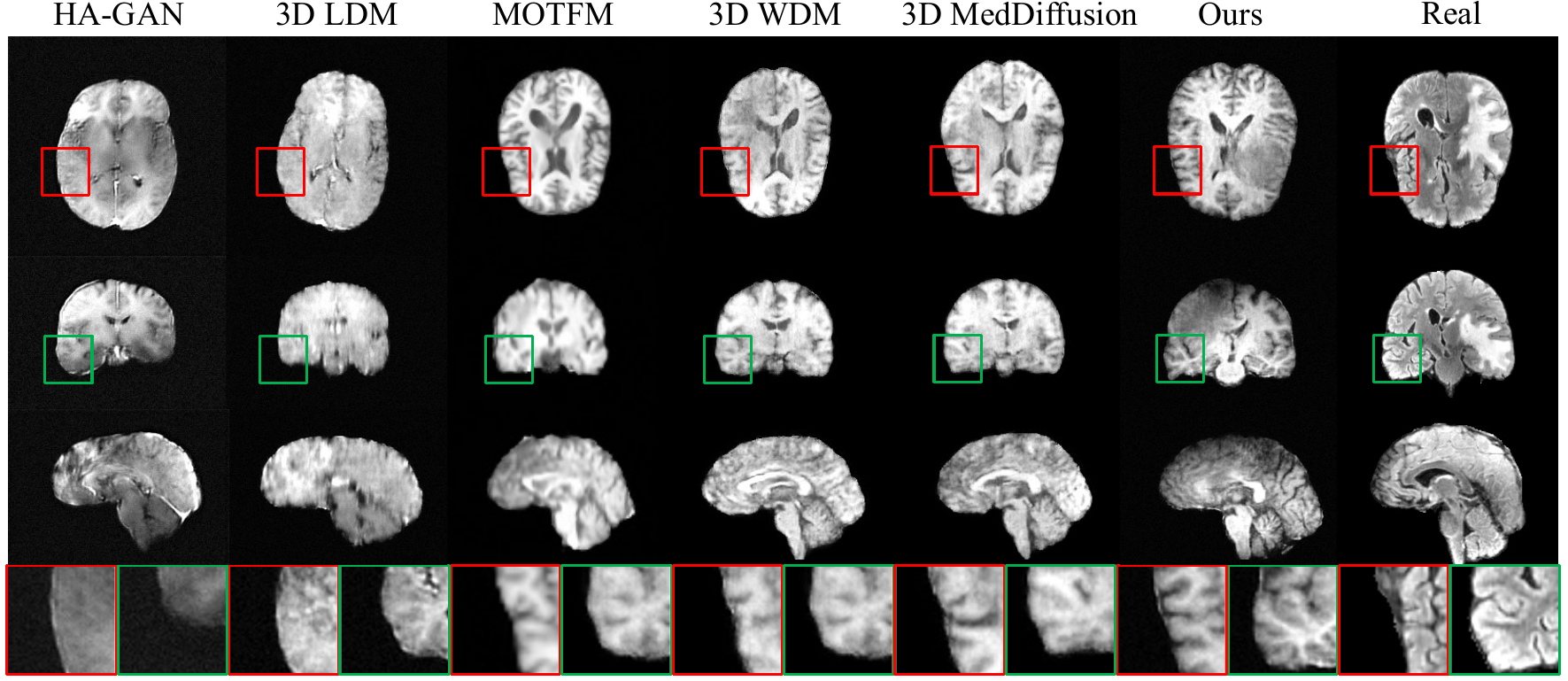}%
\caption{\textbf{Qualitative comparison on pathological brain MRI.} Each column shows the central axial, coronal, and sagittal slices of one volume. Red and green regions are enlarged for detailed comparison.
}
\label{fig:qualitative}
\end{figure*}

\subsection{Experimental Setup}

\paragraph{Datasets.}
We conduct the main generation experiments on two collections of T1-weighted brain MRI volumes. The pathological cohort comprises 1,251 volumes from the BraTS 2021 training set~\citep{baid2021brats}. 
The healthy cohort contains 958 real T1 volumes: 581 volumes from IXI~\citep{chen2021transmorph} and 252 volumes from the NIMH Healthy Research Volunteer Dataset~\citep{nugent2022nimh},  both distributed through FOMO45K~\citep{cerri2025fomo}, together with 125 manually corrected skull-stripped NFBS volumes~\citep{puccio2016nfbs}.

For both cohorts, all volumes are reoriented to the RAS coordinate system. Foreground intensities are clipped at the 0.1 and 0.99 quantiles and subsequently linearly scaled to $[-1,1]$. Each volume is standardized to $240\times240\times168$ using only zero-padding or cropping, without spatial resizing or interpolation.

\paragraph{Evaluation metrics.}
Distributional features are extracted using frozen MedicalNet 3D ResNet-50~\citep{chen2019med3d} and 3DINO ViT-L/16~\citep{xu2025_3dino} encoders. In each feature space, we compute FID~\citep{heusel2017fid} and MMD-RBF~\citep{gretton2012mmd}, where lower values indicate better distributional alignment. MedicalNet-based values are multiplied by $10^3$ for readability, while 3DINO-based values remain unscaled. Pairwise MS-SSIM~\citep{wang2003msssim} measures inter-sample similarity. We additionally report MUSIQ~\citep{ke2021musiq} and NIQE~\citep{mittal2013niqe} for perceptual quality, together with the mean 3D Tenengrad score for volumetric sharpness. MUSIQ and Tenengrad are nominally better when higher, whereas NIQE is better when lower. \textit{Detailed implementations are provided in the appendix}.

For downstream classification evaluation, we report accuracy (Acc.), balanced accuracy (Bal. Acc.), sensitivity (Sens.), and specificity (Spec.), with higher values indicating better performance. Accuracy measures the proportion of correctly classified samples, while balanced accuracy averages the class-wise recall and is more informative under class imbalance. Sensitivity measures the proportion of positive samples correctly identified, whereas specificity measures the proportion of negative samples correctly identified.

\paragraph{Implementation details.}
The Stage I structVAE, with a 64-channel bottleneck, is trained for 100 epochs. In Stage II, the generator is trained for 500 epochs on eight NVIDIA A100 GPUs using Adam with a learning rate of $10^{-5}$ and a batch size of 4. We use 32 learnable context tokens, a structure lead of $\delta=0.3$, and a structure-only warm-up probability of $p_{\mathrm{warm}}=0.15$. The weights of the 3D Sobel loss are set to $\lambda_{\mathrm{sobel}}=0.5$, respectively.
At inference, we solve the flow-matching trajectory using 100 NFEs and classifier-free guidance with a scale of 4.0. For fair comparison, all methods use the same preprocessing pipeline, MedicalNet checkpoint, metric implementations, and 1,000 generated samples per class. Each baseline retains its original training objective and sampling procedure. \textit{Further details are provided in the appendix}.

\paragraph{Comparison methods.}
We compare against five open-source 3D generators spanning complementary model families. HA-GAN is a hierarchical adversarial model that amortizes high-resolution training over subvolumes \citep{sun2022hagan}. Medical Diffusion, denoted 3D-LDM in the table, performs diffusion in the learned latent space of a 3D VQ-GAN \citep{khader2023denoising}. 3D MedDiffusion combines a Patch-Volume Autoencoder with a dual-flow latent denoiser \citep{wang2025meddiffusion}. WDM applies diffusion to an invertible 3D wavelet representation \citep{friedrich2024wdm}, whereas MOTFM uses optimal-transport flow matching and provides a three-dimensional MRI configuration \citep{yazdani2025motfm}. Each method is retrained from its official implementation.

\subsection{Main Comparison and Qualitative Analysis}

\paragraph{Quantitative comparison.}
Table~\ref{tab:main_results} shows that HA-GAN, 3D-LDM, and MOTFM exhibit large feature-distribution discrepancies and poor perceptual quality, while their high MS-SSIM values indicate limited sample diversity. The unusually high 3D Tenengrad of HA-GAN is likely caused by noise and spurious high-frequency responses rather than sharper details. 3D MedDiffusion and WDM achieve a better overall balance, but still lag behind our method across all evaluated metrics. In contrast, VoxStruct3D achieves the best overall distribution alignment, diversity, and perceptual quality across both datasets, demonstrating more realistic and structurally consistent generation.

\paragraph{Qualitative comparison.}
Figure~\ref{fig:qualitative} presents qualitative comparisons across different methods. VoxStruct3D generates volumes with clearer anatomical boundaries while better preserving global brain structures, achieving a more favorable balance between local sharpness and structural coherence. \textit{More visualizations are in the appendix.}

\begin{table}[!h]
\centering
{
\fontsize{9pt}{10.8pt}\selectfont
\rmfamily
\setlength{\tabcolsep}{1mm}
\renewcommand{\arraystretch}{1.0}

\resizebox{\columnwidth}{!}{%
\begin{tabular}{lccc}
\toprule
Method & \shortstack{Training Cost\\(A100 GPUh)} & NFE s & \shortstack{Inference\\(s/vol.)} \\
\midrule
HA-GAN & 34 & 1 & $<1$ \\
3D-LDM  & $80+32=112$ & 1000 & 102 \\
WDM & 221 & 1000 & 257 \\
MOTFM & 272 & 10 & 11 \\
3D MedDiffusion & $120+8+24=152$ & 1000 & 129 \\
\textbf{VoxStruct3D} & 258 & 100 & 63 \\
\bottomrule
\end{tabular}}
}
\caption{\textbf{Training and sampling efficiency.} NFE denotes the number of network function evaluations.}
\label{tab:efficiency}
\end{table}

\paragraph{Computational efficiency.}
Table~\ref{tab:efficiency} reports training and sampling cost on NVIDIA A100 GPUs. Inference time is averaged per volume with a batch size of 1 at a resolution of $240\times240\times168$. As a voxel-space model, VoxStruct3D incurs a relatively high training cost. Nevertheless, with only 100 NFEs, far fewer than the 1,000 used by conventional diffusion baselines, it achieves substantially better FID, MS-SSIM, and perceptual quality. Latent-space methods, by contrast, generally offer greater computational efficiency.

\begin{figure}[!t]
    \centering
    \includegraphics[width=\columnwidth]{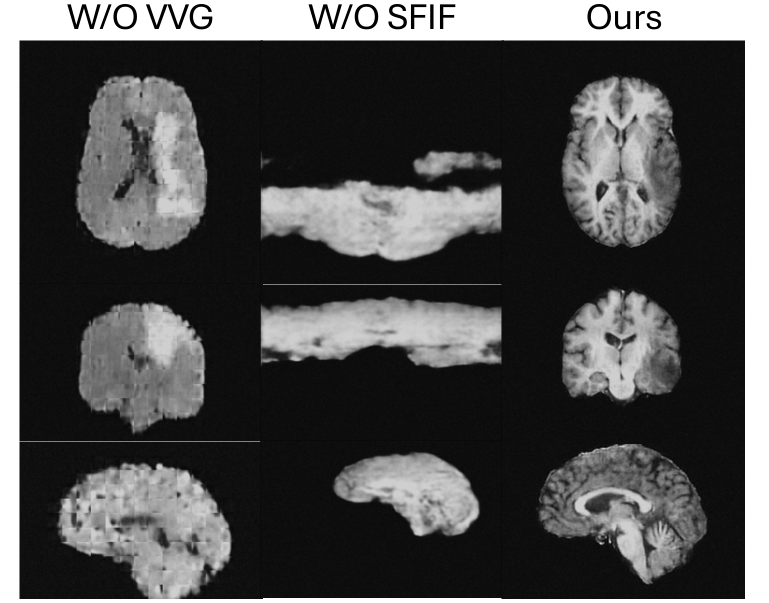}
    \caption{\textbf{Qualitative ablation of VVG and SFIF. }Without VVG, samples exhibit grid-aligned block artifacts; without SFIF, sampling can collapse toward severely malformed, anatomically invalid brain volumes.}
    \label{fig:qualitative_ablation}
\end{figure}

\subsection{Ablation Studies}

\begin{table}[!h]
\centering
{
\fontsize{9pt}{10.8pt}\selectfont
\rmfamily
\setlength{\tabcolsep}{1mm}
\renewcommand{\arraystretch}{1.0}

{%
\begin{tabular}{lcc}
\toprule
Configuration & FID$\downarrow$ & MS-SSIM$\downarrow$ \\
\midrule
Baseline & 43.41 & 0.8926 \\
+ Overlapping upsampling & 19.48 & 0.8031 \\
\ \ \ \ + Time-modulated residual refinement & 16.13 & 0.7754 \\
+ Skip fusion & 30.35 & 0.7946 \\
Full VVG & \textbf{12.89} & \textbf{0.7413} \\
\bottomrule
\end{tabular}}
}
\caption{\textbf{VVG ablation on BraTS 2021.} The indented residual-refinement row denotes the addition of time-modulated residual refinement on top of overlapping upsampling. FID is computed using 3DINO features.
}
\label{tab:VVG_ablation}
\end{table}

\paragraph{VVG ablation.}
As shown in Table~\ref{tab:VVG_ablation}, the baseline already includes SFIF but still performs poorly, indicating that structural guidance alone cannot compensate for a weak reconstruction network. Overlapping upsampling provides the largest individual improvement, reducing FID from 43.417 to 19.487 and MS-SSIM from 0.8926 to 0.8031. By allowing neighboring tokens to jointly reconstruct shared spatial regions, it effectively reduces block discontinuities and improves local coherence. Time-modulated residual refinement and skip fusion provide further gains, while combining all components yields the best FID and MS-SSIM.  The left side of Fig.~\ref{fig:qualitative_ablation} further visualizes the resulting failure modes. Removing VVG not only reduces local sharpness but also shifts boundary smoothing and edge refinement back to the Transformer, which must handle these tasks together with global token organization. Consequently, periodic block artifacts emerge and align with the token reconstruction grid.

\begin{table}[!h]
\centering
{
\fontsize{9pt}{10.8pt}\selectfont
\rmfamily
\setlength{\tabcolsep}{1mm}
\renewcommand{\arraystretch}{1.0}

{%
\begin{tabular}{cccccc}
\toprule
Str. flow & StructVAE    & Str. leading & PA-RoPE      & Asy. mask   & FID$\downarrow$ \\
\midrule
$\times$     & $\times$     & $\times$      & $\times$     & $\times$     & 18.28 \\
$\checkmark$ & $\times$     & $\times$      & $\times$     & $\times$     & 25.47 \\
$\checkmark$ & $\checkmark$ & $\times$      & $\times$     & $\times$     & 18.09 \\
$\checkmark$ & $\checkmark$ & $\checkmark$  & $\times$     & $\times$     & 14.92 \\
$\checkmark$ & $\checkmark$ & $\times$      & $\checkmark$ & $\times$     & 16.89 \\
$\checkmark$ & $\checkmark$ & $\times$      & $\times$     & $\checkmark$ & 17.38 \\
$\checkmark$ & $\checkmark$ & $\checkmark$  & $\checkmark$ & $\checkmark$ & \textbf{12.89}\\
\bottomrule
\end{tabular}}

\caption{\textbf{SFIF ablation on BraTS 2021.} Checkmarks denote enabled components; crosses denote disabled ones.  FID is computed using 3DINO features.}
\label{tab:sfif_ablation}
}
\end{table}

\paragraph{SFIF ablation.}
Table~\ref{tab:sfif_ablation} evaluates SFIF on top of the full VVG baseline. Directly introducing structure flow on raw 3DINO features degrades FID, indicating increased optimization difficulty from the information-rich structural target. structVAE compresses these features into compact anatomy-focused tokens and largely removes this degradation. The structure-leading clock lets structural information be recovered earlier to guide image generation, while PA-RoPE aligns interactions between unequal 3D token grids and the asymmetric attention mask prevents structure tokens from copying image appearance. Combining all components achieves the best FID of 12.89. As shown in the middle of Fig.~\ref{fig:qualitative_ablation},removing SFIF leads to a different failure mode: at high noise levels, the pixel stream contains little reliable global anatomical information, and without the progressively clarified structure state, sampling may enter an invalid anatomical configuration that later local refinement cannot correct.

\subsubsection{Downstream classification.}
To test whether generated volumes preserve condition-discriminative anatomy and provide useful supervision for downstream learning, we evaluate three 3D CNN classifiers. AD-DL-Image follows the Conv5\_FC3 subject-level network of Wen et al., whereas AD-DL-Patch follows their Conv4\_FC3 patch-level network~\citep{wen2020convolutional}.  Dilated-3D adapts the dilated 3D convolutional design of Liu et al.~\citep{liu2021enhancing}. The synthetic training set contains 1,000 pathological and 1,000 healthy volumes. Each classifier follows its official training settings and is trained for 12 epochs using (i) the real training set, (ii) the synthetic set, or (iii) their union.Each setting is run five times with different random seeds on the same untouched real test set of 188 pathological and 144 healthy subjects, with results reported as mean${\pm}$standard deviation.
\textit{Further details are provided in the appendix}.

\begin{table}[!h]
\centering
{%
\fontsize{9pt}{10.8pt}\selectfont
\rmfamily
\setlength{\tabcolsep}{1mm}
\renewcommand{\arraystretch}{1.0}
{%
\begin{tabular}{lcccc}
\toprule
Classifier
& Acc.$\uparrow$
& Bal. Acc.$\uparrow$
& Sens.$\uparrow$
& Spec.$\uparrow$ \\
\midrule

\multicolumn{5}{l}{\textit{Real training set}} \\

AD-DL-Image
& $92.8{\pm}0.8$ & $91.8{\pm}0.9$ & $99.4{\pm}0.5$ & $84.2{\pm}1.7$ \\

AD-DL-Patch
& $94.5{\pm}0.7$ & $94.6{\pm}0.6$ & $94.0{\pm}1.1$ & $\mathbf{95.1{\pm}0.9}$ \\

Dilated-3D
& $95.7{\pm}0.5$ & $95.3{\pm}0.6$ & $98.8{\pm}0.5$ & $91.7{\pm}1.1$ \\

\midrule
\multicolumn{5}{l}{\textit{Synthetic training set}} \\

AD-DL-Image
& $89.8{\pm}1.0$ & $89.5{\pm}1.0$ & $92.1{\pm}1.4$ & $86.8{\pm}1.4$ \\

AD-DL-Patch
& $93.7{\pm}0.7$ & $92.9{\pm}0.8$ & $\mathbf{98.7{\pm}0.6}$ & $87.1{\pm}1.4$ \\

Dilated-3D
& $93.6{\pm}0.8$ & $94.2{\pm}0.7$ & $89.9{\pm}1.4$ & $\mathbf{98.5{\pm}0.8}$ \\

\midrule
\multicolumn{5}{l}{\textit{Real + synthetic training sets}} \\

AD-DL-Image
& $\mathbf{95.6{\pm}0.6}$ & $\mathbf{95.0{\pm}0.6}$ & $\mathbf{99.7{\pm}0.3}$ & $\mathbf{90.3{\pm}1.1}$ \\

AD-DL-Patch
& $\mathbf{95.8{\pm}0.6}$ & $\mathbf{95.6{\pm}0.6}$ & $97.5{\pm}0.7$ & $93.8{\pm}1.1$ \\

Dilated-3D
& $\mathbf{97.8{\pm}0.4}$ & $\mathbf{97.6{\pm}0.4}$ & $\mathbf{99.4{\pm}0.5}$ & $95.8{\pm}0.9$ \\

\bottomrule
\end{tabular}}}

\caption{
\textbf{Downstream classification.} Results using real, synthetic, and combined training sets, evaluated on the same untouched real test set. Bold indicates the best mean across settings.
}
\label{tab:downstream_classification}
\end{table}
As shown in Table~\ref{tab:downstream_classification}, synthetic-only training trails real-only training by only 1.97 accuracy points on average, indicating that the generated volumes retain condition-discriminative information. Combining real and synthetic data further improves average accuracy and balanced accuracy by 2.09 and 2.19 points, respectively, showing that synthetic volumes provide complementary training variation.

\FloatBarrier

\section{Conclusion}
We presented VoxStruct3D, a voxel-space framework for high-fidelity 3D MRI generation. VVG couples neighboring token predictions through overlapping volumetric decoding, residual refinement, and skip fusion, while SFIF provides progressively clarified anatomical guidance through a compact 3DINO-derived structure stream. Experiments on pathological and healthy brain MRI show strong distributional and perceptual quality, and the component ablations verify the complementary contributions of VVG and SFIF. Downstream classification further evaluates whether the generated volumes retain condition-related information.

\bibliography{aaai2027}

\appendix
\clearpage

\twocolumn[
\begingroup
\let\Large\LARGE
\section{Appendix}
\label{app:appendix}
\endgroup
]

\subsection{Hyperparameter Settings}
\label{app:hyperparameters}

Table~\ref{tab:implementation_hyperparameters} summarizes the
implementation settings used in the two stages and at inference.

\begin{table*}[t]
\centering
{
\small
\setlength{\tabcolsep}{4pt}
\renewcommand{\arraystretch}{1.30}
\begin{tabular}{@{}p{0.37\textwidth}@{\hspace{1em}}p{0.56\textwidth}@{}}
\toprule
Setting & Value \\
\midrule
Structure lead $\delta$ & $0.3$ \\
Structure-only warm-up probability $p_{\mathrm{warm}}$ & $0.15$ \\
Stage-I StructVAE bottleneck width & $64$ \\
StructVAE input/output code width & $1024/1024$ \\
StructVAE token grid & $7\times7\times7=343$ \\
Stage-I loss weights & MSE $1.0$; cosine $1.0$; KL $10^{-7}$; code mean $0.01$; code standard deviation $0.01$ \\
Stage-II image/structure velocity-loss weights & Image $1.0$; structure $1.0$ \\
CFG label-drop probability & $0.1$ \\
CFG inference scale & $4.0$ \\
CFG interval & Global time $(0.0,1.0)$ with strict endpoint inequalities \\
ODE solver & Heun; Euler at terminal and structure-terminal crossings \\
Default sampling budget & 50 integration steps; 100 NFEs \\
Image volume size & $1\times240\times240\times168$ \\
Image token grid & $10\times10\times7=700$ \\
Structure token grid & $7\times7\times7=343$ \\
DiT & 12 blocks; width $1024$; 16 heads; MLP expansion ratio $4$ \\
Context tokens & 32; inserted before the fourth block \\
Stage-II Sobel-loss weight & $0.5$ \\
Sobel time weight & $w(t_x)=\left[(t_x-0.3)/(1-0.3)\right]_+^{1.5}$, clipped to $[0,1]$ \\
Main-generator parameters & $324{,}214{,}883$ total; $323{,}498{,}083$ trainable \\
StructVAE parameters & $104{,}120{,}448$ \\
3DINO teacher & ViT-L/16; 24 blocks; width $1024$; 16 heads; final normalized patch tokens \\
Stage-II optimizer & Adam; $\beta=(0.9,0.95)$; weight decay $0.0$ \\
Stage-II EMA & $0.9999$ and $0.9996$, updated after each optimizer step \\
\bottomrule
\end{tabular}
}
\caption{Hyperparameter and implementation settings.}
\label{tab:implementation_hyperparameters}
\end{table*}

\subsection{Metrics}
\label{app:metric}

\subsubsection{Volumetric Distribution and Diversity Metrics}
\label{app:metric_volumetric}

For MedicalNet-based FID and inter-sample MS-SSIM, we use the volumetric
evaluators released with WDM \citep{friedrich2024wdm} and recompute all scores
on our real and generated volumes under a unified evaluation protocol.

\paragraph{MedicalNet FID.}
Following the WDM MedicalNet-FID evaluator, we use the frozen MedicalNet 3D ResNet-50
\texttt{resnet\_50\_23dataset} checkpoint \citep{chen2019med3d}. Features are
taken after the fourth residual stage and reduced by
\texttt{AdaptiveAvgPool3d(1)}, yielding one $2048$-dimensional descriptor per
volume. The empirical means and covariance matrices of the real and generated
descriptors are denoted by $(\boldsymbol{\mu}_r,\boldsymbol{\Sigma}_r)$ and
$(\boldsymbol{\mu}_g,\boldsymbol{\Sigma}_g)$, respectively. We compute
\begin{equation}
    \operatorname{FID}=
    \lVert\boldsymbol{\mu}_r-\boldsymbol{\mu}_g\rVert_2^2+
    \operatorname{Tr}\!\left(
    \boldsymbol{\Sigma}_r+\boldsymbol{\Sigma}_g-
    2(\boldsymbol{\Sigma}_r\boldsymbol{\Sigma}_g)^{1/2}
    \right).
    \label{eq:medicalnet_fid}
\end{equation}
As in 3D-WDM, the matrix square root is evaluated with
\texttt{scipy.linalg.sqrtm}; if the covariance product is numerically singular,
$10^{-6}$ is added to both covariance diagonals. The same frozen descriptors
are used for MedicalNet MMD-RBF. MedicalNet-based FID and MMD-RBF values are
multiplied by $10^3$ only when displayed in the tables.

\paragraph{3DINO feature extraction.}
For the 3DINO feature space, each single-channel volume is passed through the
frozen high-resolution 3DINO ViT-L/16
\citep{xu2025_3dino}. We use the model's standard image-level inference output:
the class token after the final Transformer block and final LayerNorm
(\texttt{x\_norm\_clstoken}). This gives one $1024$-dimensional descriptor per
volume for FID and MMD-RBF, and the corresponding reported values are not
rescaled. This evaluation descriptor is distinct from the $343$ final
normalized patch tokens (\texttt{x\_norm\_patchtokens}) used by the Stage-I
structural teacher; the latter retain the $7\times7\times7$ spatial token grid
and are not used as the volume-level distribution descriptor.

\paragraph{Inter-sample MS-SSIM.}
Following the WDM MS-SSIM evaluator, we use MONAI Generative's
\texttt{MultiScaleSSIMMetric} with \texttt{spatial\_dims=3},
\texttt{data\_range=1.0}, and \texttt{kernel\_size=7}. The commonly
preprocessed volumes are mapped to $[0,1]$ before this calculation. For an
evaluated set of $N$ volumes, we compute the full-volume score for every pair
with distinct sample identities and report their arithmetic mean. Although
the reference implementation enumerates the $N(N-1)$ ordered pairs, symmetry
makes this mean identical to averaging the $N(N-1)/2$ unordered pairs. Lower
inter-sample MS-SSIM indicates lower average structural redundancy and hence
greater pairwise diversity. Together with FID and MMD-RBF, inter-sample
MS-SSIM characterizes both distributional fidelity and sample diversity.

For NIQE and MUSIQ, each NIfTI volume is first reoriented to the closest
RAS orientation. A brain mask $M$ is obtained by applying Otsu thresholding
to the finite-valued voxels, followed by one 3D morphological opening,
retention of the largest connected component, closing, hole filling, and two
dilations. Let $V_M=\{V(v):v\in M\}$ denote the voxel intensities inside the
mask, and let $P_q(\cdot)$ denote the $q$-th percentile. Intensity
normalization uses only these foreground voxels:
\begin{equation}
    \begin{aligned}
        l&=P_{0.5}(V_M), & h&=P_{99.5}(V_M),\\
        \widehat V&=\operatorname{clip}\left(\frac{V-l}{h-l},0,1\right).
    \end{aligned}
    \label{eq:metric_normalization}
\end{equation}
All voxels outside $M$ are subsequently set to zero.

We extract slices along the axial, coronal, and sagittal axes. For each axis
$a$, slices are selected at relative positions $r\in\{0.2,0.5,0.8\}$ within
the brain-containing range:
\begin{equation}
    k_{a,r}=\operatorname{round}\left(
    k_a^{\min}+r\left(k_a^{\max}-k_a^{\min}\right)\right).
    \label{eq:metric_slice_selection}
\end{equation}
Thus, each volume contributes at most nine slices. Each slice is cropped using
the 3D brain bounding box, masked again, and zero-padded directly to
$256\times256$. No resizing, interpolation, or other spatial resampling is
applied, so the original pixel grid and brain scale are preserved. The single
MRI channel is then repeated three times, $I_{\mathrm{RGB}}=[I,I,I]$. For either
2D metric
$m\in\{\mathrm{NIQE},\mathrm{MUSIQ}\}$, aggregation is performed first over
slices and then over volumes:
\begin{equation}
    m_v=\frac{1}{N_{s,v}}\sum_{s=1}^{N_{s,v}}m(I_{v,s}),
    \qquad
    m_{\mathrm{set}}=\frac{1}{N_v}\sum_{v=1}^{N_v}m_v.
    \label{eq:metric_aggregation}
\end{equation}
This gives every volume equal weight.

\subsubsection{NIQE}
\label{app:metric_niqe}

NIQE \citep{mittal2013niqe} is computed with the default \texttt{pyiqa}
implementation. It first forms mean-subtracted contrast-normalized (MSCN)
coefficients
\begin{equation}
    \operatorname{MSCN}(i,j)=
    \frac{I(i,j)-\mu(i,j)}{\sigma(i,j)+C}.
    \label{eq:niqe_mscn}
\end{equation}
Here, $I$ is the grayscale slice represented on $[0,255]$; $\mu(i,j)$ and
$\sigma(i,j)$ are its local weighted mean and standard deviation, respectively,
computed using the default normalized $7\times7$ Gaussian window with standard
deviation $7/6$ and replicate padding; and $C=1$ is a stabilizing constant.
The implementation then
extracts natural-scene-statistics features from the MSCN coefficients and
products of neighboring coefficients. Let
$(\boldsymbol{\mu}_r,\boldsymbol{\Sigma}_r)$ denote the pretrained reference
feature distribution and
$(\boldsymbol{\mu}_I,\boldsymbol{\Sigma}_I)$ the distribution estimated from
the evaluated image. The NIQE score is
\begin{equation}
    d_{\mathrm{NIQE}}=
    \sqrt{
    (\boldsymbol{\mu}_r-\boldsymbol{\mu}_I)^\top
    \left(\frac{\boldsymbol{\Sigma}_r+\boldsymbol{\Sigma}_I}{2}\right)^{-1}
    (\boldsymbol{\mu}_r-\boldsymbol{\mu}_I)}.
    \label{eq:niqe_distance}
\end{equation}
A lower score indicates a smaller deviation from the NIQE reference model.

\subsubsection{MUSIQ}
\label{app:metric_musiq}

MUSIQ \citep{ke2021musiq} is evaluated using the default \texttt{pyiqa}
configuration and MUSIQ weights pretrained on KonIQ-10k. The model receives the
three-channel $256\times256$ slices in $[0,1]$ and predicts
\begin{equation}
    q_{\mathrm{MUSIQ}}=f_{\theta}(I_{\mathrm{RGB}}),
    \label{eq:musiq_score}
\end{equation}
where $f_{\theta}$ is the pretrained multi-scale image-quality Transformer.
The slice predictions are aggregated using Eq.~\eqref{eq:metric_aggregation}.
A higher score indicates higher predicted perceptual quality.

\subsubsection{3D Tenengrad}
\label{app:metric_3d_tenengrad}

We measure volumetric sharpness using 3D Tenengrad, defined as the
foreground-averaged squared 3D Sobel-gradient energy. After the selected volume
normalization and center padding/cropping, the foreground mask is defined as
$M=\{v:V(v)>0.05\}$. If it contains fewer than 64 voxels, we instead use
$M=\{v:|V(v)|>10^{-6}\}$. Three Sobel responses are computed along the voxel
axes with nearest-boundary padding:
\begin{equation}
    \begin{aligned}
        G_x&=\operatorname{Sobel}_x(V),&
        G_y&=\operatorname{Sobel}_y(V),\\
        G_z&=\operatorname{Sobel}_z(V).
    \end{aligned}
    \label{eq:3d_sobel}
\end{equation}
The per-voxel squared gradient energy and volume-level score are
\begin{equation}
    \begin{aligned}
        E_{\mathrm{3D}}(v)
        &=G_x(v)^2+G_y(v)^2+G_z(v)^2,\\
        T_{\mathrm{3D}}
        &=\frac{1}{|M|}\sum_{v\in M}E_{\mathrm{3D}}(v).
    \end{aligned}
    \label{eq:3d_tenengrad}
\end{equation}
Dataset-level scores are obtained by averaging $T_{\mathrm{3D}}$ equally across
volumes. Higher values indicate stronger volumetric edge content. We report 3D
Tenengrad alongside perceptual and distributional metrics as a complementary
measure of volumetric sharpness. However, noise and artifacts can also elevate
the squared-gradient response, so an excessively high Tenengrad value may
reflect spurious high-frequency content rather than improved image quality and
should not be interpreted as uniformly better in isolation.

\subsection{Classification Data and Preprocessing}

\paragraph{Data partition.}

\begin{table*}[t]
\centering
{
\small
\setlength{\tabcolsep}{5pt}
\begin{tabular}{llrrrrr}
\toprule
Data type & Source / condition &
Total & Train & Validation & Test & Split ratio \\
\midrule
\multirow{6}{*}{Real}
& BraTS 2021 (pathological) & 1,251 & 876 & 187 & 188
& 70.0/15.0/15.0 \\
& IXI (healthy)            & 581   & 407 & 87  & 87
& 70.1/15.0/15.0 \\
& NIMH (healthy)           & 252   & 176 & 38  & 38
& 69.8/15.1/15.1 \\
& NFBS (healthy)           & 125   & 87  & 19  & 19
& 69.6/15.2/15.2 \\
\cmidrule(lr){2-7}
& Healthy subtotal         & 958   & 670 & 144 & 144
& 69.9/15.0/15.0 \\
& All real volumes         & 2,209 & 1,546 & 331 & 332
& 70.0/15.0/15.0 \\
\midrule
\multirow{3}{*}{Synthetic}
& VoxStruct3D pathological & 1,000 & 1,000 & 0 & 0 & Training only \\
& VoxStruct3D healthy      & 1,000 & 1,000 & 0 & 0 & Training only \\
\cmidrule(lr){2-7}
& All synthetic volumes    & 2,000 & 2,000 & 0 & 0 & Training only \\
\bottomrule
\end{tabular}
}
\caption{
Volume counts for the subject-level data partition and use of synthetic
volumes. Within each real-data source, subjects were assigned to training,
validation, and test subsets using an approximately $70\%$/$15\%$/$15\%$
split. The split was fixed across all classifiers, and all scans, volumes, and
patches from a given subject were kept in the same subset. VoxStruct3D
generated 1,000 pathological and 1,000 healthy volumes, which were used
exclusively for classifier training and never for validation or testing.
}
\label{tab:data_split}
\end{table*}

The real data were partitioned once at the subject level using a fixed,
source-stratified manifest. Within each source, subjects were assigned to the
training, validation, and test splits in an approximately 70\%/15\%/15\%
ratio. The counts reported in Table~\ref{tab:data_split} refer to volumes,
yielding 1,546 training volumes, 331 validation volumes, and 332 test volumes.
The test set comprised 188 pathological volumes from BraTS 2021 and 144
healthy volumes, including 87 from IXI, 38 from NIMH, and 19 from NFBS\@. All
scans and volumes from a given subject, together with all patches extracted
from them, were assigned to the same split; thus, no subject contributed data
to more than one split. The same subject-level manifest was used for all three
downstream classifiers.

VoxStruct3D uses the pathological/healthy cohort label as its conditioning
variable, matching the binary downstream task.
After generator training, we sampled 1,000 pathological and 1,000
healthy synthetic volumes. Synthetic volumes were used exclusively
as classifier training data and were never included in the
validation or test sets. The real validation set was used for model
selection, whereas all final classification results were computed
on the same held-out real test set. Accordingly, the real-only,
synthetic-only, and real-plus-synthetic settings differed only in
their classifier training data and were evaluated using identical
real validation and test volumes.

\paragraph{Data processing.}
\label{app:classification_data_processing}

All three classifiers use the same fixed, source-stratified, subject-level
training, validation, and test manifest. Consequently, every volume and every
patch from a given subject remains in the same split. Real and synthetic
volumes undergo the same processing pipeline. The preprocessing is designed to
reduce
site-, scanner-, and acquisition-specific variation, encouraging the
classifiers to rely on pathology-related anatomy rather than multicenter
imaging characteristics. Each NIfTI volume is loaded
as a 32-bit floating-point array, and non-finite values are replaced with
zero. We define the foreground as the set of positive voxels and crop the
volume to its three-dimensional foreground bounding box. The box is expanded
along each axis by 5\% of the foreground extent, with a minimum margin of two
voxels, and is clipped to the image boundaries. Restricting the field of view
to the foreground reduces variation arising from background padding and
differences in acquisition coverage.

Intensity normalization is performed independently for each volume using only
its foreground voxels. In the default rank-normalization setting, foreground
intensities are sorted and the empirical rank $r_i$ of voxel $i$ is mapped to
\begin{equation}
    p_i=\frac{r_i+0.5}{N}, \qquad
    \widetilde{x}_i=
    \frac{\operatorname{clip}\!\left(\Phi^{-1}(p_i),-3,3\right)}{3},
    \label{eq:classification_rank_normalization}
\end{equation}
where $N$ is the number of foreground voxels and $\Phi^{-1}$ is the inverse
cumulative distribution function of the standard normal distribution.
Background voxels remain zero and do not contribute to the normalization
statistics. This operation maps the foreground intensities into $[-1,1]$
while reducing between-volume differences in global intensity scale
and marginal intensity distributions that may be associated with acquisition
sites, scanners, and protocols.

After cropping and normalization, each volume is resized to a standardized
$128\times128\times128$ input grid using trilinear interpolation. A
singleton channel dimension is then added, giving an input tensor of shape
$1\times128\times128\times128$. During training only, intensity augmentation
multiplies the volume by a factor sampled uniformly from $[0.90,1.10]$ and
adds an offset sampled uniformly from $[-0.05,0.05]$. Gaussian noise with
standard deviation $0.015$ is also added with probability $0.3$.
These perturbations approximate residual differences in intensity gain,
offset, and noise level, thereby discouraging the classifiers from relying on
site-specific intensity signatures. No augmentation is used for validation
or testing.

The image-level classifiers receive the complete processed volume. For the
patch-level classifier, $64\times64\times64$ subvolumes are extracted after
the common preprocessing steps. One patch per subject is used during
training, whereas eight fixed patches are evaluated for each subject during
validation and testing. If $\mathbf{z}_{s,k}$ denotes the logits of patch
$k$ from subject $s$, the subject-level prediction is obtained from
\begin{equation}
    \overline{\mathbf{z}}_s
    =\frac{1}{K}\sum_{k=1}^{K}\mathbf{z}_{s,k},
    \qquad K=8.
    \label{eq:classification_patch_aggregation}
\end{equation}
We then select the class with the largest averaged logit. Thus, all
reported classification metrics use subjects, rather than individual
patches, as the unit of evaluation.

\subsection{Sensitivity to the Structure Lead}
\label{app:structure_lead_ablation}

We conduct a controlled sweep over the structure lead
$\delta\in\{0,0.15,0.30,0.45,0.60\}$ on BraTS 2021. All configurations use
the full architecture, including both VVG and SFIF, and share the same training
schedule, random-seed protocol, 50-step solver, 100-NFE budget, and
classifier-free-guidance scale of 4.0; only $\delta$ is changed. When $\delta=0$,
the image and structure streams follow synchronized flow-matching clocks.
Positive values move the structure stream ahead of the image stream, with
larger values imposing a stronger structure lead. We report 3DINO FID and
MS-SSIM, consistent with the main ablation studies. This controlled sweep
evaluates synchronized, intermediate, and larger structure leads.

\begin{table}[!ht]
\centering
{
\small
\setlength{\tabcolsep}{10pt}
\renewcommand{\arraystretch}{1.15}
\begin{tabular}{ccc}
\toprule
$\delta$ & FID $\downarrow$ & MS-SSIM $\downarrow$ \\
\midrule
$0$    & 16.23 & 0.8153 \\  
$0.15$ & 14.56 & 0.7821 \\  
$0.30$ & 12.89 & 0.7413 \\ 
$0.45$ & 13.24 & 0.7789 \\  
$0.60$ & 14.87 & 0.7656 \\  
\bottomrule
\end{tabular}
}
\caption{Sensitivity to the structure lead $\delta$ on BraTS 2021. All
settings other than $\delta$ are held fixed. FID is computed using 3DINO
features. The default setting used in the main experiments is
$\delta=0.30$.}
\label{tab:structure_lead_sensitivity}
\end{table}
\FloatBarrier

The sweep shows that moderate nonzero structure leads
($\delta\in\{0.15,0.30,0.45\}$) perform well, with all three settings improving
both metrics over synchronized generation ($\delta=0$). Among the evaluated
values, $\delta=0.30$ attains the best FID (12.89) and MS-SSIM (0.7413),
compared with 16.23 and 0.8153 at $\delta=0$; we therefore select
$\delta=0.30$ as a representative default for the main experiments.

\subsection{Stage-I Architecture}
\label{app:stage1_architecture}

\paragraph{Structural teacher.}
The Stage-I input is a single-channel volume resized by trilinear
interpolation to $112\times112\times112$. A frozen 3DINO ViT-L/16 with 24
Transformer blocks, width 1024, and 16 attention heads divides the volume into
a $7\times7\times7$ patch grid. We retain its final normalized patch tokens
and discard the class token, producing
$\mathbf{r}\in\mathbb{R}^{B\times343\times1024}$.

\paragraph{StructVAE encoder.}
A linear projection $1024\!\rightarrow\!1024$ first maps each descriptor, and
a fixed three-dimensional sine--cosine positional embedding is added without
changing the token grid. The encoder then applies four pre-LayerNorm
Transformer blocks. Each block contains eight-head self-attention and a
GELU MLP with an expansion ratio of 4, i.e., hidden dimensions
$1024\!\rightarrow\!4096\!\rightarrow\!1024$, together with residual
connections. The attention and MLP dropout rates are both set to zero. A final
LayerNorm and a linear head $1024\!\rightarrow\!128$ predict a 64-dimensional
posterior mean and a 64-dimensional log-variance for every token. The resulting latent tensor
has shape $B\times343\times64$.

\paragraph{Structure-code projection and decoder.}
A linear layer $64\!\rightarrow\!1024$ projects each latent token to the
structure-code space, yielding
$\mathbf{s}\in\mathbb{R}^{B\times343\times1024}$. The decoder begins with a
$1024\!\rightarrow\!1024$ projection, adds the same fixed 3D positional
embedding, and uses another four Transformer blocks with the same width,
eight-head attention, pre-LayerNorm, GELU MLP expansion ratio, and residual
layout as the encoder. A final LayerNorm and $1024\!\rightarrow\!1024$
reconstruction head recover the 3DINO descriptor grid. Thus, StructVAE compresses channel
width from 1024 to 64 while preserving all 343 spatial tokens. It contains
$104{,}120{,}448$ parameters, excluding the frozen 3DINO teacher.

\paragraph{StructVAE validation results.}

Table~\ref{tab:stage1_reconstruction} reports the Stage-I StructVAE results on
the validation set. StructVAE achieves a reconstruction MSE of 0.04261 and a
reconstruction cosine similarity of 0.98173, corresponding to a cosine loss of
0.01827 and indicating accurate recovery of the 3DINO descriptors. The latent
codes have a mean of $-0.01181$ and a standard deviation of 0.99027, close to
their regularization targets of zero and one, respectively.

\begin{table}[!h]
\centering
{
\small
\setlength{\tabcolsep}{8pt}
\renewcommand{\arraystretch}{1.15}
\begin{tabular}{lc}
\toprule
Metric & Value \\
\midrule
Reconstruction MSE $\downarrow$ & \textbf{0.04261} \\
Cosine loss $\downarrow$ & \textbf{0.01827} \\
Reconstruction cosine similarity $\uparrow$ & \textbf{0.98173} \\
Code mean $\rightarrow 0$ & \textbf{$-0.01181$} \\
Code standard deviation $\rightarrow 1$ & \textbf{0.99027} \\
\bottomrule
\end{tabular}
}
\caption{Stage-I StructVAE validation results.}
\label{tab:stage1_reconstruction}
\end{table}
\FloatBarrier

\subsection{Stage-II Architecture}
\label{app:stage2_architecture}

\paragraph{Timestep conditioning in convolutional blocks.}
The convolutional encoder and overlapping decoder use GroupNorm-based FiLM
modulation to inject the image timestep. Given a convolutional feature
$\mathbf{h}$, the image timestep is first embedded as
\begin{equation}
    \mathbf{c}_{t_x}=\operatorname{TimeEmbed}(t_x),
    \label{eq:conv_time_embedding}
\end{equation}
and a linear projection of the SiLU-activated embedding predicts channel-wise
scale and shift parameters:
\begin{equation}
    [\boldsymbol{\gamma}_{t_x},\boldsymbol{\beta}_{t_x}]
    =\operatorname{Linear}\!\left(
    \operatorname{SiLU}(\mathbf{c}_{t_x})\right).
    \label{eq:conv_film_parameters}
\end{equation}
These parameters are broadcast over the three spatial dimensions and applied
to the GroupNorm output as
\begin{equation}
    \widetilde{\mathbf{h}}
    =(1+\boldsymbol{\gamma}_{t_x})\odot
    \operatorname{GroupNorm}(\mathbf{h})
    +\boldsymbol{\beta}_{t_x}.
    \label{eq:conv_adagn}
\end{equation}
This GroupNorm-based FiLM operation is an AdaGN-style timestep modulation and
allows convolutional features to adapt to the image noise level $t_x$.

\paragraph{Image-token encoder.}
The image stream receives
$\mathbf{x}\in\mathbb{R}^{B\times1\times240\times240\times168}$. Three
timestep-conditioned 3D convolutions successively map
\[
\begin{aligned}
1\times240\times240\times168
&\xrightarrow[\;1\to16\;]{k=s=2}
16\times120\times120\times84\\
&\xrightarrow[\;16\to256\;]{k=s=3}
256\times40\times40\times28\\
&\xrightarrow[\;256\to1024\;]{k=s=4}
1024\times10\times10\times7.
\end{aligned}
\]
A SiLU activation is applied between the second and third convolutional stages.
The final feature grid is flattened into 700 image tokens of width 1024, while
the 16-channel half-resolution feature is retained for the decoder skip
connection.

\paragraph{Shared dual-stream DiT.}
The 700 image tokens and 343 structure tokens of width 1024 are processed by 12
shared Transformer blocks of width 1024. Each block uses 16-head self-attention
(head width 64), QKV bias, QK normalization, RMSNorm with
$\epsilon=10^{-6}$, a SwiGLU feed-forward network with an expansion ratio of
4, and AdaLN-Zero shift, scale, and gate modulation. The image and structure
streams have separate timestep embedders and share a class embedding. Thirty-two
context tokens are inserted before the fourth block; after insertion,
the sequence order is
\[
[\text{context};\ \text{image};\ \text{structure}].
\]
Image and context queries can attend to all tokens, whereas structure queries
can attend only to context-token and structure-token keys. For spatial
attention, 3D rotary embeddings place the $10\times10\times7$ image grid and the
$7\times7\times7$ structure grid in a common continuous coordinate system;
context tokens use identity rotary embeddings. After the shared stack,
an independent AdaLN-modulated linear head maps the structure tokens to the
1024-dimensional clean structure endpoint.

\paragraph{Overlapping image decoder.}
The image tokens are reshaped to
$1024\times10\times10\times7$. A timestep-conditioned
$\operatorname{ConvTranspose3D}(k=8,s=4,p=2)$ maps channels
$1024\rightarrow256$ and produces $40\times40\times28$, followed by two
timestep-conditioned residual 3D convolutional refinement blocks. A second
$\operatorname{ConvTranspose3D}(k=6,s=3,p=2,o=1)$ maps
$256\rightarrow32$ and produces $120\times120\times84$, again followed
by two refinement blocks. This tensor is concatenated with the retained
16-channel encoder feature. A timestep-conditioned $3\times3\times3$
convolution maps the 48 concatenated channels to 32, followed by two more
refinement blocks.
Finally,
$\operatorname{ConvTranspose3D}(k=4,s=2,p=1)$ maps
$32\rightarrow1$ and reconstructs
$240\times240\times168$. Because the first two decoder kernels are larger
than their strides, neighboring token projections overlap before the final
voxel prediction.

\subsection{Additional Visualizations}
\label{app:more_visualizations}

\begin{figure*}[t]
    \centering
    \includegraphics[width=0.8\textwidth]{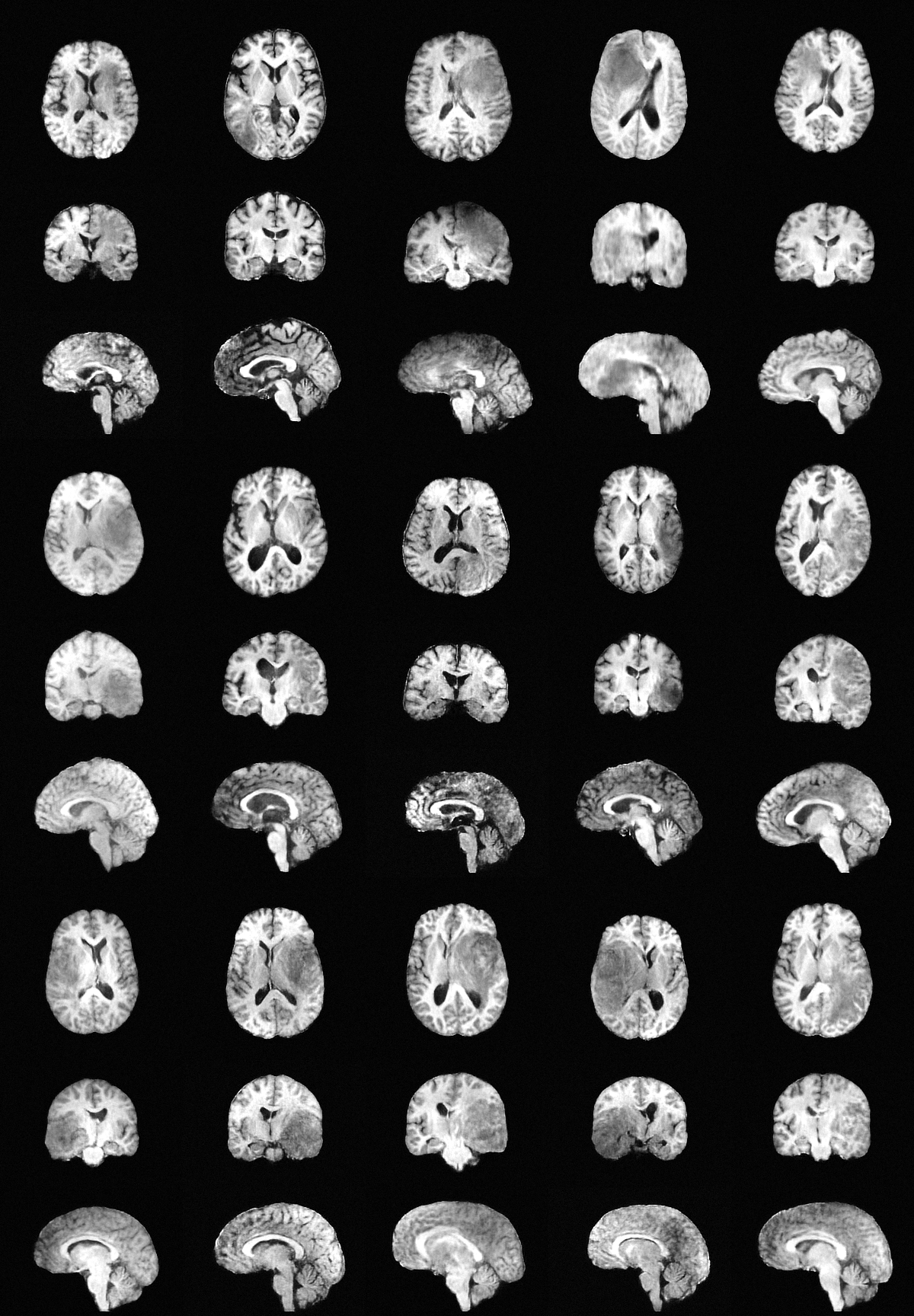}
    \caption{Additional pathological T1 MRI volumes produced by VoxStruct3D.
    }
    \label{fig:more_visualizations1}
\end{figure*}

\begin{figure*}[t]
    \centering
    \includegraphics[width=0.8\textwidth]{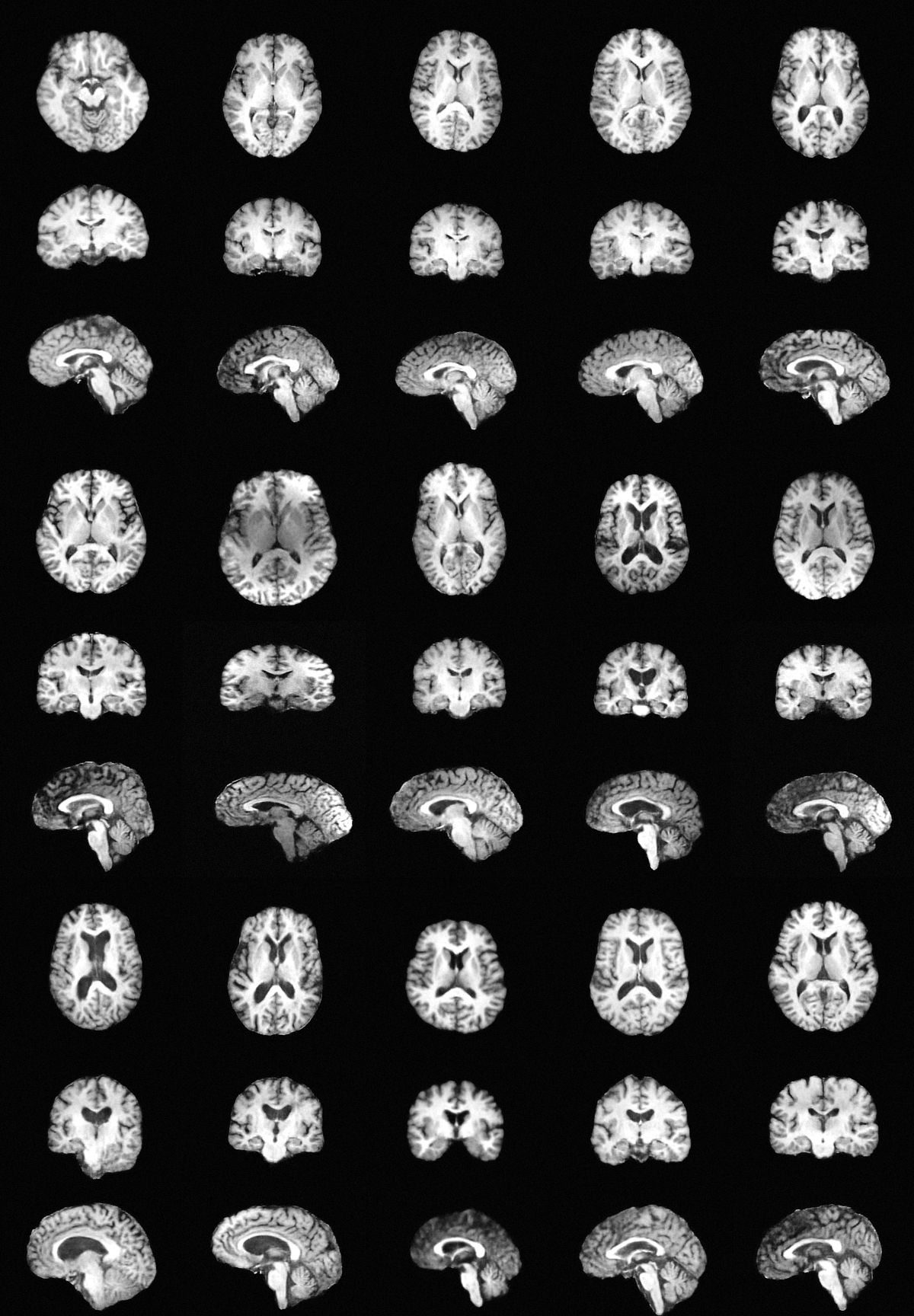}
    \caption{Additional healthy T1 MRI volumes produced by VoxStruct3D.}
    \label{fig:more_visualizations2}
\end{figure*}


\end{document}